\documentclass[letterpaper]{article} 
\usepackage{aaai2026}  
\usepackage{times}  
\usepackage{helvet}  
\usepackage{courier}  
\usepackage[hyphens]{url}  
\usepackage{graphicx} 
\usepackage{natbib}  
\usepackage{caption} 
\usepackage{algorithm}
\usepackage{algorithmic}

\usepackage{newfloat}
\usepackage{listings}
\DeclareCaptionStyle{ruled}{labelfont=normalfont,labelsep=colon,strut=off} 
\floatstyle{ruled}
\newfloat{listing}{tb}{lst}{}
\floatname{listing}{Listing}
\usepackage{multirow, makecell}
\usepackage{array}
\usepackage{cutwin}
\usepackage{marvosym}
\usepackage{booktabs}
\usepackage{amsmath, amssymb}
\usepackage{pifont}

\newcolumntype{L}[1]{>{\raggedright\arraybackslash}p{#1}}   
\newcolumntype{C}[1]{>{\centering\arraybackslash}p{#1}}     
\newcolumntype{R}[1]{>{\raggedleft\arraybackslash}p{#1}}    

\title{YOLO-IOD: Towards Real Time Incremental Object Detection}
\author{
        Shizhou Zhang\textsuperscript{\rm 1}, Xueqiang Lv\textsuperscript{\rm 1}, Yinghui Xing\textsuperscript{\rm 1}\thanks{Corresponding Author: xyh\_7491@nwpu.edu.cn}, Qirui Wu\textsuperscript{\rm 1}, Di Xu\textsuperscript{\rm 2}, Chen Zhao\textsuperscript{\rm 1}, Yanning Zhang\textsuperscript{\rm 1}
}
\affiliations{
    \textsuperscript{\rm 1}Northwestern Polytechnical University, China ~~~~
    
    \textsuperscript{\rm 2}	Huawei, China
}

\begin{document}

\maketitle

\begin{abstract}
Current methods for incremental object detection (IOD) primarily rely on Faster R-CNN or DETR series detectors; however, these approaches do not accommodate the real-time YOLO detection frameworks.
In this paper, we first identify three primary types of knowledge conflicts that contribute to catastrophic forgetting in YOLO-based incremental detectors: foreground-background confusion, parameter interference, and misaligned knowledge distillation.
Subsequently, we introduce \textbf{YOLO-IOD}, a real-time Incremental Object Detection (IOD) framework that is constructed upon the pretrained YOLO-World model, facilitating incremental learning via a stage-wise parameter-efficient fine-tuning process. 
Specifically, YOLO-IOD encompasses three principal components: 1) Conflict-Aware Pseudo-Label Refinement (CPR), which mitigates the foreground-background confusion by leveraging the confidence levels of pseudo labels and identifying potential objects relevant to future tasks. 2) Importance-based Kernel Selection (IKS), which identifies and updates the pivotal convolution kernels pertinent to the current task during the current learning stage. 3) Cross-Stage Asymmetric Knowledge Distillation (CAKD), which addresses the misaligned knowledge distillation conflict by transmitting the features of the student target detector through the detection heads of both the previous and current teacher detectors, thereby facilitating asymmetric distillation between existing and newly introduced categories.
We further introduce \textbf{LoCo COCO}, a more realistic benchmark that eliminates data leakage across stages.
Experiments on both conventional and LoCo COCO benchmarks show that YOLO-IOD achieves superior performance with minimal forgetting. 


\begin{figure}[t!]
    \centering
    \includegraphics[width=0.98\linewidth, height=0.90\linewidth]{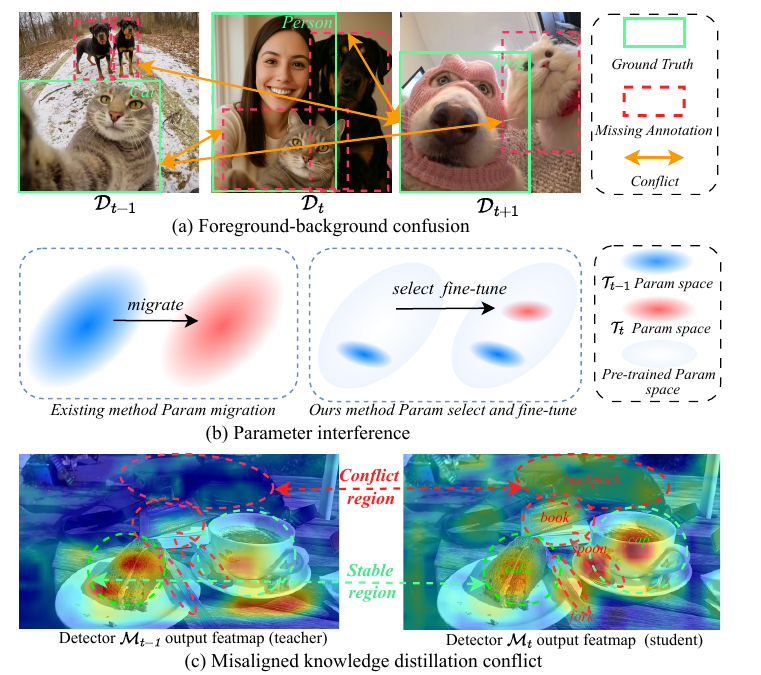}
    \caption{Illustration of the three major \textit{knowledge conflicts}: (a) Foreground-background confusion, (b) Parameter interference and (c) Misaligned knowledge distillation conflict.
    }
    \label{fig:motivation}
\end{figure}

\end{abstract}


\section{Introduction}




The goal of Incremental Object Detection (IOD) is to consistently acquire new object classes over learning of successive tasks, while retaining the knowledge of those learned in the past.
While recent works~\cite{mo2024bridge, liu2023continual} have made significant progress in IOD, most existing methods are built upon detectors such as Faster R-CNN~\cite{ren2015faster} or DETR~\cite{carion2020end}. 
When these developed methods are applied to real-time YOLO series detectors~\cite{redmon2016you}, they usually suffer significant drops in generalization performance and struggle to preserve prior knowledge of previous classes, mainly due to knowledge conflicts across tasks.


In this work, we firstly identify the underlying causes of forgetting in YOLO-based incremental detectors as three major types of \textit{knowledge conflicts} as shown in Fig.~\ref{fig:motivation}: 
\textit{1) Foreground-background confusion}, where unannotated objects of previous and future tasks are misclassified as background during training. 
This problem is critical for YOLO detectors, which rely on aggressive data augmentation techniques like Mosaic and MixUp that assume accurate annotations. In IOD settings, noise from pseudo labels is amplified by these augmentations, compromising model performance.
\textit{2) Parameter interference}. This issue arises because different tasks frequently depend on intersecting parameter subsets within the model. Updates for new tasks can alter these shared parameters, thereby potentially disrupting previously acquired representations and leading to catastrophic forgetting of earlier tasks.
\textit{3) Misaligned knowledge distillation conflict}, where the teacher and student models are optimized for mismatched class distributions, violating the core assumption in standard knowledge distillation that both models share consistent learning objectives. 
YOLO-series detectors, known for making dense predictions across spatial grids, are significantly affected by this problem.


To overcome above challenges, we propose YOLO-IOD, a real-time IOD framework that builds upon the pre-trained YOLO-World~\cite{cheng2024yolo} model and performs parameter efficient fine-tuning during each incremental learning stage. 
YOLO-IOD comprises three main modules designed to address the identified knowledge conflicts: 
1) Conflict-Aware Pseudo-Label Refinement (CPR), which alleviates foreground-background confusion by incorporating two strategies: Enhanced Pseudo-label Loss, which improves the reliability of supervision from pseudo-labels by weighting predictions based on their confidence and uncertainty and Clustered Unknown Pseudo Labeling, which identifies potential objects from future tasks via performing open-vocabulary object detection and feature-space clustering.
2) Importance-based Kernel Selection (IKS), which mitigates parameter interference by selecting and fine-tuning only important task-relevant convolution kernels based on Fisher-based differential importance estimation.
and 3) Cross-Stage Asymmetric Knowledge Distillation (CAKD), which resolves misaligned knowledge distillation conflict by passing cross-stage student features through detection heads of both old and current teacher detectors, enabling asymmetric distillation across old and new categories.


Moreover, current IOD benchmarks are not tailored for practical applications as they divide categories arbitrarily, disregarding natural co-occurrence of classes, and permit images to repeat across incremental stages. 
In practical situations, certain categories such as cars and pedestrians frequently occur together, whereas others, like cars and boats, do not. Images overlapping between stages must be strictly prevented to avoid data leakage.
This raises questions about whether the performance observed on current benchmarks can effectively translate to real-world applications, particularly for recent IOD methods that depend on pseudo labels.
Therefore, we introduce LoCo COCO, a novel benchmark designed to eliminate inter-stage image overlap and adhere to category co-occurrence statistics, offering a more equitable and realistic evaluation benchmark for IOD.

Our contributions can be summarized as follows:

\begin{itemize}
    \item We introduce YOLO-IOD, an integrated and real-time IOD framework, and pinpoint three causes of forgetting: foreground-background confusion, parameter interference, and misaligned knowledge distillation conflict.
    
    \item YOLO-IOD incorporates three innovative modules aimed at mitigating forgetting, with particular emphasis on the dual-teacher CAKD module. This module addresses the challenge of misaligned knowledge distillation by channeling the target student detector's features through the detection heads of both the former and current teacher detectors.
    
    \item We introduce LoCo COCO, a practical benchmark designed to remove image overlap between stages and consider category co-occurrence, allowing for a more equitable assessment of incremental object detection.
    
    \item Extensive experiments on conventional COCO and LoCo COCO benchmarks under multiple settings show that our method consistently achieves state-of-the-art performance while maintaining real-time inference speed. 

\end{itemize}

\section{Related Works}

\subsection{Incremental Learning}


Incremental learning enables models to acquire new information without forgetting prior knowledge~\cite{schlimmer1986incremental, li2017learning}. Approaches include \textit{rehearsal-based methods}, which replay or generate samples from previous tasks~\cite{rebuffi2017icarl, lopez2017gradient, shin2017continual}; \textit{regularization-based methods}, which constrain important parameters to maintain stability~\cite{kirkpatrick2017overcoming, gda_iod, yang2023continual, repre}; \textit{architecture-based methods}, which assign task-specific subnetworks~\cite{von2019continual, rypesc2024divide}; and \textit{knowledge distillation}, which transfers information from past tasks via teacher-student learning~\cite{tao2020topology, chen2023dynamic, asadi2023prototype}. 
Recently, \textit{continual learning with foundation models} have also attracted many interests, including approaches such as L2P~\cite{wang2022learning} CODA-Prompt~\cite{smith2023coda} and VPT-NSP$^2$~\cite{lu2024visual}.

\begin{figure*}
    \centering
    \includegraphics[width=0.98\linewidth]{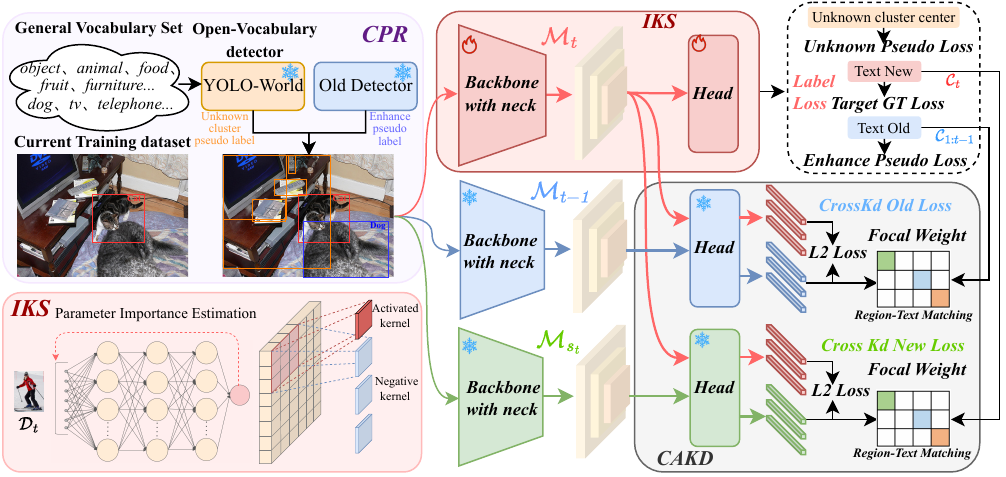}
    \caption{Overall architecture of YOLO-IOD. Our method integrates three components to address knowledge conflicts in IOD: 1) Conflict-aware Pseudo-label Refinement (CPR); 2) Importance-based Kernel Selection (IKS); and 3) Cross-Stage Asymmetric Knowledge Distillation (CAKD).
}
    \label{fig:overallframework}
\end{figure*}

\subsection{Incremental Object Detection}
Compared to classification, IOD involves more complex scenarios and greater challenges due to the requirement of both localizing and classifying the objects. 
In IOD, distillation is particularly effective due to the rich feature representations and multi-level supervision signals available in detection frameworks. 
ERD~\cite{feng2022overcoming} introduces elastic response distillation that adaptively transfers knowledge. 
BPF~\cite{mo2024bridge} bridges past-future knowledge through dual teacher models, CL-DETR~\cite{liu2023continual} applies distillation within DETR architectures, emphasizing reliable predictions from old models. 
Nevertheless, these methodologies tackle the issue of knowledge distillation by selecting only those old task outputs that do not coincide with new labels to serve as the distillation targets. 
This strategy is unsuitable for YOLO-style dense predictions. Moreover, this approach can only distill partial knowledge from teacher models, failing to fully resolve the underlying conflict. 
We propose CAKD which provides cross-stage asymmetric knowledge distillation by suppressing the responses of unrelated features with the detection heads.

\section{Preliminaries and Benchmark}

\subsection{Problem Definition}
In IOD, tasks arrive sequentially in an ordered sequence: $\mathcal{T} = \{ \mathcal{T}_1, \mathcal{T}_2, \ldots, \mathcal{T}_t, \ldots, \mathcal{T}_n \}$. 
Each task $\mathcal{T}_t$ aims to learn a specific set of object categories $\mathcal{C}_t$, and $\quad \mathcal{C}_i \cap \mathcal{C}_j = \emptyset, \; \forall i \ne j$. 
For each task $\mathcal{T}_t$, a dataset is provided: $\mathcal{D}_t = \{ (\mathcal{X}_t^i, \mathcal{Y}_t^i) \}_{i=1}^{N_t}$,
where $\mathcal{X}_t^i$ denotes the $i$-th image and $\mathcal{Y}_t^i$ is its corresponding annotation. Importantly, the dataset $\mathcal{D}_t$ is annotated only for categories in $\mathcal{C}_t$.
That is, objects belonging to $\mathcal{C}_{1:t-1} \cup \mathcal{C}_{t+1:n}$ are not annotated, even if presented in the image. 
During the training of task $\mathcal{T}_t$, only $\mathcal{D}_t$ is accessible. 
The goal of IOD is to train an object detector $\mathcal{M}_t$ at each stage, such that it can correctly detect objects from both the current task's classes $\mathcal{C}_t$ and all seen classes $\mathcal{C}_{1:t-1}$.







\subsection{YOLO-World}

Open-vocabulary object detection (OVOD) enables detectors to recognize arbitrary categories guided by textual input. 
Recent methods such as GLIP~\cite{Li_2022_CVPR}, Grounding DINO~\cite{liu2024grounding}, and YOLO-World~\cite{cheng2024yolo} learn region-text alignment on large-scale vision-language data. 
Our method builds upon the pre-trained YOLO-World model. 
Images are processed by visual encoder $f_v$ and category texts by text encoder $f_t$, with features fused using RepVL-PAN: $\mathbf{V}, \mathbf{P} = f_{\text{RepVL}}(f_v(I), f_t(T))$ where $\mathbf{V} = \{\mathbf{e}_k\}_{k=1}^K$ are region-level visual embeddings and $\mathbf{P} = \{\mathbf{p}_j\}_{j=1}^C$ are text prototypes. Classification scores are computed as $s_{k,j} = \eta \cdot \langle \mathrm{Norm}(\mathbf{e}_k), \mathrm{Norm}(\mathbf{p}_j) \rangle + \zeta$.

\subsection{LoCo COCO Benchmark}
In prior IOD benchmark, each stage $t$ typically selects all images that contain objects of categories $C_t$ from the full dataset. 
Real-world images frequently contain objects from various categories, meaning that a single image will be used across several training stages. 
Based on the statistics, each image appears in an average of \textit{1.84} stages in the 20+20 4 stage setting. 
This overlap challenges the foundational premise of continual learning and artificially inflates the effectiveness of pseudo-labeling methods by allowing detectors to generate pseudo-labels on reused training images.

To address this issue, we introduce a new data partitioning protocol named as Low Co-occurrence COCO (LoCo COCO). 
We first construct a category co-occurrence matrix $\mathbf{A} \in \mathbb{R}^{N \times N}$, where $\mathbf{A}_{ij}$ denotes the number of images in which categories $c_i$ and $c_j$ co-occur. This matrix defines an undirected weighted graph $\mathcal{G}=(\mathcal{V}, \mathcal{E})$, with nodes for categories and edge weights given by $\mathbf{A}_{ij}$. 
We then perform graph clustering on $\mathcal{G}$ to divide the category set $\mathcal{C} = \{c_1, c_2, \ldots, c_n\}$ into $n$ disjoint subsets for $n$ stages. This ensures that categories frequently co-occur are assigned to the same task, minimizing inter-task image overlap. 
After the above category partitioning, there still exists a portion of overlap images that contain categories from multiple stages. 
For each overlapping image $I$ that spans multiple candidate category sets,  we randomly assign image $I$ to one of the candidate tasks. 
This strategy ensures that each image appears in only one stage, diminishing any data leakage across stages.
Our proposed LoCo COCO benchmark aligns better with the real-world IOD scenario and eliminates the evaluation bias in pseudo-labeling based IOD methods.
\section{Method}
\subsection{Overall Framework}

The proposed YOLO-IOD utilizes the pretrained YOLO-World as its foundational model. 
As illustrated in Fig.~\ref{fig:overallframework}, YOLO-IOD performs incremental learning through stage-wise partial fine-tuning, where parameter updates are guided by importance-based kernel selection. 
The overall architecture consists of three components. 
In the upper-left part, \textit{Conflict-aware Pseudo-label Refinement (CPR)} generates high-quality old task supervision through an enhanced pseudo-label loss to mitigate foreground-background confusion from previous tasks, while addressing potential foreground-background confusion on future tasks through Clustered Unknown Pseudo Labeling. 
In the lower-left part of Fig.~\ref{fig:overallframework}, \textit{Importance-based Kernel Selection (IKS)} determines which convolutional kernels to be updated at each incremental stage, preserving critical knowledge from previous tasks while allowing partial fine-tuning to new categories. 
Finally, \textit{Cross-Stage Asymmetric Knowledge Distillation (CAKD)} implements two distinct knowledge distillation losses.
The first involves comparing the old detector $\mathcal{M}_{t-1}$ with the target detector $\mathcal{M}_{t}$ by feeding their respective dense features into the head of the old detector $\mathcal{M}_{t-1}$. 
The second involves comparing the current detector $\mathcal{M}_{s_t}$ (the detector trained on current stage dataset $D_t$ annotated for categories $\mathcal{C}_t$) with the target detector $\mathcal{M}_{t}$ by utilizing the dense features within the current detector's head.




\begin{table*}[ht]
\centering

\begin{tabular}{L{1cm}|L{1.85cm}C{1.65cm}C{2.7cm}|C{0.55cm}C{0.55cm}C{0.55cm}C{0.55cm}C{0.55cm}C{0.55cm}|C{0.95cm}C{0.95cm}}
\toprule
Setting & Method & References & Baseline & $ AP $ & $AP_{50}$ & $AP_{75}$ & $AP_S$ & $AP_M$ & $AP_L$ & $AbsGap$ & $RelGap$ \\
\midrule

\multirow{5}{*}{\shortstack{Upper\\Bound}}
& Joint Train & NeurIPS 15 & Faster R-CNN & 40.2 & 61.6 & 43.5 & 23.5 & 43.9 & 52.2 & - & -\\
& Joint Train & ICLR 21 & Deformable DETR & 47.0 & 66.1 & 50.9 & - & - & - & - & -\\
& Joint Train & CVPR 24 & YOLO-World & 54.5 & 71.3 & 59.7 & 37.2 & 60.1 & 70.1 & - & -\\
& Joint Train & CVPR 22 & GLIP-T & 55.2 & - & - & - & - & - & - & -\\
& Joint Train & ECCV 24 & Grounding Dino & 57.2 & - & - & - & - & - & - & -\\

\midrule

\multirow{10}{*}{40+40} 
& BPF & ECCV 24 & Faster R-CNN & 34.4 & 54.3 & 37.3  & -    & -    & -    & 5.8& 14.4\% \\
& $\mathrm{RGR}^*$ & CVPR 25 & Faster R-CNN & 35.6 & 56.0 & 38.2  & -    & -    & -    & 4.6 & 11.4\% \\
\cmidrule(l){2-12}
& $\mathrm{CL\textbf{-}DETR}^*$ & CVPR 23 & Deformable DETR & 42.0 & 60.1 & 45.9 & 24.0 & 45.3 & 55.6 & 5.0 & 10.6\%\\
& DCA & AAAI 25 & Deformable DETR & 42.8 & 58.4 & -  & - & - & - & 4.2 & 8.9\%\\
& $\mathrm{SDDGR}^*$ & CVPR 24 & Deformable DETR & 43.0 & 62.1 & 47.1 & 24.9 & 46.9 & 57.0 & 4.0 & 8.5\%\\
\cmidrule(l){2-12}

& TLR & AAAI 24 & GLIP-T & 40.4 & 57.4 & 43.9 & 23.3 & 44.7 & 54.5 & 14.8 & 26.8\%\\

& GCD & AAAI 25 & Grounding Dino & 45.7 & 62.9 & 49.7 & 28.4 & 49.3 & 60.0 & 11.5 & 20.1\%\\

& ERD & CVPR 22 & YOLO-Word(X) & 49.9 & 67.0  & 54.7  & 33.5  & 54.8  &  64.8 &   4.6 & 8.4\%\\

& $\mathrm{RGR}^*$ & CVPR 25 & YOLO-Word(X) & 51.5  & 68.1  &  56.5 &  34.1 & 57.0  & 66.9  &   3.0 &  5.5\%\\

& \textbf{YOLO-IOD}& Ours & YOLO-Word(X) & \textbf{53.0} & \textbf{69.7} & \textbf{58.1} & \textbf{36.1} & \textbf{58.6} & \textbf{67.0} & \textbf{1.5} & \textbf{2.7\%}\\

\midrule

\multirow{10}{*}{70+10} 
& BPF & ECCV 24 & Faster R-CNN & 36.2 & 56.8 & 38.9  & -    & -    & -    & 4& 9.9\% \\
& $\mathrm{RGR}^*$ & CVPR 25 & Faster R-CNN & 36.6 & 56.6 & 39.6  & -    & -    & -    & 3.6 & 8.9\% \\
\cmidrule(l){2-12}
& $\mathrm{CL\textbf{-}DETR}^*$ & CVPR 23 & Deformable DETR & 40.4 & 58.0 & 43.9 & 23.8 & 43.6 & 53.5 & 6.6 & 14.0\%\\
& $\mathrm{SDDGR}^*$ & CVPR 24 & Deformable DETR & 40.9 & 59.5 & 44.8 & 23.9 & 44.7 & 54.0 & 6.1 & 13.0\%\\
& DCA & AAAI 25 & Deformable DETR & 41.3 & 59.2 & - & - & - & - & 5.7 & 12.1\%\\
\cmidrule(l){2-12}

& TLR & AAAI 24 & GLIP-T & 42.9 & 59.2 & 45.2 & 24.3 & 45.1 & 54.1 & 12.3 & 22.2\%\\

& GCD & AAAI 25 & Grounding Dino & 46.7 & 63.9 & 50.8 & 29.7 & 49.9 & 61.6 & 8.5 & 14.8\%\\

& ERD & CVPR 22 & YOLO-Word(X) & 45.6 & 62.0  & 50.1  &  29.9 & 50.6  &  60.6 &   8.9 & 16.3\%\\

& $\mathrm{RGR}^*$ & CVPR 25 & YOLO-Word(X) & 49.1  &  64.9 &  53.5 &  31.2 & 54.1  & 64.8  &  5.4& 9.9\%\\

& \textbf{YOLO-IOD}& Ours & YOLO-Word(X) & \textbf{52.4} & \textbf{68.9} & \textbf{57.4} & \textbf{35.9} & \textbf{58.0} & \textbf{65.6} & \textbf{2.1} & \textbf{3.9\%} \\
\bottomrule
\end{tabular}
\caption{Comparison with methods in single step settings. Replay-based methods are marked with~$\ast$. Results are from original papers except those YOLO-World based methods which we reproduced using publicly available code. Best results in bold.}
\label{tab:single_stage}
\end{table*}

\subsection{Conflict-Aware Pseudo-Label Refinement}

CPR comprises two components. The Enhanced Pseudo-label Loss leverages confidence-aware supervision along with entropy regularization for better use of pseudo-labels. Meanwhile, Clustered Unknown Pseudo Labeling incorporates open-vocabulary object detection and feature clustering to offer consistent supervision for objects in future tasks that are not annotated. 

\textbf{Enhanced Pseudo-label Loss.} 
Traditional pseudo-labeling methods in IOD rely on confidence thresholds to select reliable pseudo labels, which introduces confidence-based selection bias. This bias causes low-confidence categories to be gradually ignored during training. In addition, pseudo labels above the threshold are treated uniformly without considering their actual reliability, resulting in inconsistent supervision signals. 

To address this, we propose an Enhanced Pseudo-label Loss that treats the confidence score $s$ of each pseudo label as a soft supervision target, integrating confidence-aware weighting with entropy regularization:
\begin{equation}
\mathcal{L}_{pseudo}^{cls} = -|s - p_t|^{\gamma} \log(p_t) + \lambda \cdot (1 - s)^{\delta} \cdot H(\hat{y}),
\label{eq:enhance_pseudo}
\end{equation}
where $p_t$ represents the predicted probability for the pseudo-labeled class, and $H(\hat{y})$ denotes the entropy of the predicted category distribution. 
The first term achieves confidence aligned supervision using a focal-style scheme, while the second term applies adaptive entropy regularization that scales inversely with confidence scores. 
This design takes full advantage of pseudo labels across wide range of confidence scores. 
Low-confidence pseudo labels provide soft supervision and are further regularized by the second term in Eq.~(\ref{eq:enhance_pseudo}), which retain uncertainty in the predictions, while high-confidence labels contribute more stable supervision, proportional to their reliability.

\textbf{Clustered Unknown Pseudo Labeling.} 
For the missing annotations of future task categories, we propose Clustered Unknown Pseudo Labeling method. We construct a General Vocabulary Set $V_{gen}$ comprising 500 common object categories and 50 abstract super-categories summarized by a large language model.
During each incremental stage, we apply YOLO-World using $V_{gen}$ to identify all foreground objects excluding those with ground-truth annotations.
Let $\mathcal{F}$ denote the set of these unannotated foreground predictions by YOLO-World, $\mathcal{C}_{\mathcal{F}} \subseteq V_{\text{gen}}$ represent the set of category labels of $\mathcal{F}$. 
To convert these predictions into stable unknown supervision while minimizing conflicts with future tasks, we perform frequency-weighted K-Means clustering on the text feature representations $f_t(\cdot)$ for $\mathcal{C}_{\mathcal{F}}$, where the weight of each category corresponds to its occurrence frequency in $\mathcal{F}$.
The resulting cluster centroids define a set of unknown super categories $\mathcal{U} = \{u_1, u_2, \ldots, u_K\}$. We then replace each label in $\mathcal{F}$ with its assigned super-category in $\mathcal{U}$, and substitute their text embeddings with the corresponding cluster centroids during training.
This approach transforms the knowledge conflict arising from unannotated future task categories into a process of discovering and learning new classes from the unknown super-categories.


\subsection{Importance-based Kernel Selection}
We utilize an importance based parameter selection mechanism to alleviate the parameter conflict in YOLO-IOD.
Specifically, we select and fine-tune only important convolution kernels in each incremental task, minimizing interference with the overall parameter distribution.
To avoid disrupting critical knowledge from previous tasks, we compute differential importance by subtracting historical importance from current task-specific importance. 

\textbf{Parameter Importance Estimation.} We adopt Fisher Information to quantify parameter importance but define it at the granularity of convolution kernels to preserve inductive structure and avoid prohibitive storage costs as tasks increase.
Given a convolution kernel \( \mathbf{w}^k = \{ w_j^k \}_{j=1}^{d_k} \) consisting of \( d_k \) scalar parameters, its Fisher-based importance can be computed as :
\begin{equation}
\mathbf{I}_t(\mathbf{w}^k) = \sum_{j=1}^{d_k} \left( \frac{1}{N_t} \sum_{n=1}^{N_t} \left( \frac{\partial \log p(y_n \mid x_n; \theta)}{\partial w_j^k} \right)^2 \right),
\end{equation}
where $(x_n, y_n)$ are training samples from task $\mathcal{T}_t$, and \( \theta \) denotes the model parameters. To avoid interference with previously learned tasks, we compute the differential importance, where \( \rho \) denotes a weighting factor:
\begin{equation}
\Delta \mathbf{I}_t(\mathbf{w}^k) = \mathbf{I}_t(\mathbf{w}^k) - \rho \sum_{i=1}^{t-1} \mathbf{I}_i(\mathbf{w}^k).
\end{equation}
We then select the top-\(\mathcal{K}\) kernels ranked by \( \Delta \mathbf{I}_t(\mathbf{w}^k) \) for fine-tuning in task $\mathcal{T}_t$, while keeping the rest frozen.

\subsection{Cross-Stage Asymmetric Knowledge Distillation}

As illustrated in Fig.~\ref{fig:overallframework}, our CAKD module employs a dual-teacher framework, where the target detector $\mathcal{M}_{t}$ serves as the student. 
The first teacher is the old detector $\mathcal{M}_{t-1}$, which is dedicated to the previously learned categories $\mathcal{C}_{1:t-1}$ and primarily attends to the foreground features of $\mathcal{C}_{1:t-1}$, with its head suppressing responses to unrelated features. 
The second teacher is the current detector $\mathcal{M}_{s_t}$, whose features focus on the current stage classes while suppressing features of other categories. 
This cross-stage asymmetric knowledge distillation design enables the target detector $\mathcal{M}_t$ to avoid misaligned supervision and feature interference between tasks, while maximally distilling and integrating knowledge across both old and new categories.

The distillation process operates by passing student neck features $\mathbf{F}_{\text{student}}^{\text{neck}}$ to teachers' detection heads, generating cross stage post-head features. 
The detection head comprises classification and regression components: the regression head outputs bounding box positions for each anchor, while the classification head produces image encodings subjected to Region-Text Matching with text embeddings to yield classification logits.

We apply distillation loss globally across the feature map. 
To suppress noisy or background regions and focus on the most informative and reliable predictions, we introduce a focal weight $w_{\text{focal}}(p) = \max_j \mathrm{logit}_{\text{teacher}}(p, j)$ for each spatial location $p$, emphasizing foreground-likely areas based on teacher's maximal confidence.

The classification distillation loss measures the $L_2$ distance between teacher and student image-region encodings at each location, weighted by the focal factor:
\begin{equation}
\mathcal{L}_{\text{cls\_kd}} = \sum_{p} \| \mathbf{E}_{\text{teacher}}(p) - \mathbf{E}_{\text{student\_cross}}(p) \|_2^2 \cdot w_{\text{focal}}(p),
\end{equation}
where $\mathbf{E}_{\text{teacher}}(p)$ and $\mathbf{E}_{\text{student\_cross}}(p)$ denote the region-level feature embeddings from the teacher and student models, respectively. Similarly, the regression distillation loss operates over all positions in the feature map, with the background-suppressing effect achieved via $w_{\text{focal}}(p)$:
\begin{equation}
\mathcal{L}_{\text{reg\_kd}} = \sum_{p} \mathcal{L}_{\text{IoU}}(\text{B}_{\text{tea}}(p), \\ \text{B}_{\text{stu\_cross}}(p))
\cdot w_{\text{focal}}(p),
\end{equation}
where $\text{B}_{\text{stu\_cross}}(p)$ and $\text{B}_{\text{tea}}(p)$ are the bounding boxes from the student and teacher models at location $p$. The overall distillation objective is the weighted sum of these two parts:
\begin{equation}
\mathcal{L}_{\text{CAKD}} = \alpha\, \mathcal{L}_{\text{cls\_kd}} + \beta\, \mathcal{L}_{\text{reg\_kd}}.
\end{equation}


\section{Experiments}

\begin{table*}[ht]
\centering

\begin{tabular}{l|c|C{0.4cm}C{0.4cm}C{1.05cm}|C{0.4cm}C{0.4cm}C{1.05cm}|C{0.4cm}C{0.4cm}C{1.05cm}|C{0.4cm}C{0.4cm}C{1.05cm}}
\toprule

\multirow{2.3}{*}{Method} & \multirow{2.3}{*}{Detector} & \multicolumn{3}{c|}{40-10} & \multicolumn{3}{c|}{40-20} & \multicolumn{3}{c|}{20-20} & \multicolumn{3}{c}{10-10} \\

& & $\mathcal{T}_3$ & $\mathcal{T}_5$ & $RelGap$ &$\mathcal{T}_2$ & $\mathcal{T}_3$ & $RelGap$&$\mathcal{T}_2$ & $\mathcal{T}_4$ & $RelGap$ &$\mathcal{T}_4$  & $\mathcal{T}_8$ & $RelGap$ \\
\midrule
$\mathrm{CL\textbf{-}DETR}^*$  & Deformable DETR & -    & 28.1 & 40.2\% & -    & 35.3 & 24.8\% & -    & 34.2    &  27.2\%   & - & 24.4 & 48.1\% \\
$\mathrm{SDDGR}^*$      & Deformable DETR & 40.6 & 36.8 & 21.7\% & 42.5 & 41.1 & 12.5\% & -    & -    & -    & -    & - & - \\
DCA        & Deformable DETR & 41.1 & 37.2 & 20.8\% & 42.7 & 40.3 & 14.2\% & -    & -    & -    & -    & - & - \\
\midrule
TLR        & GLIP-T          & -    & 30.2 & 45.2\% & -    & 37.3 & 32.4\% & -    & -    & -    & -    & -    & -    \\
GCD        & Grounding Dino  & -    & 40.2 & 29.7\% & -    & 44.0 & 23.0\% & -    & -    & -    & -    & -    & -    \\
ERD        & YOLO-Word(X)    & 47.3 & 37.9 & 30.4\% & 50.0 & 46.4 & 14.8\% & 55.1 & 44.1 & 19.1\% & 46.9 & 32.0 & 41.2\% \\ 
$\mathrm{RGR}^*$        & YOLO-Word(X)    & 48.2 & 44.8 & 17.8\% & 51.0 & 48.6 & 10.8\% & 56.2 & 48.1 & 11.7\% & 52.3 & 43.4 & 20.3\%\\
\textbf{Ours} & YOLO-Word(X) & \textbf{51.8} & \textbf{50.6} & \textbf{7.1\%} & \textbf{52.9} & \textbf{51.9} & \textbf{4.8\%} & \textbf{58.3} & \textbf{51.7} & \textbf{5.1\%} & \textbf{56.8} & \textbf{49.7} & \textbf{8.8\%} \\
\bottomrule
\end{tabular}
\caption{Incremental results (AP, \%) under multi-step settings. Results for 20-20, 10-10, and all YOLO-World based methods are reproduced using publicly available code. Methods using example-replay are marked with $\ast$. Best results in bold.}
\label{tab:multi_step_results}
\end{table*}

\begin{table*}
\centering
\begin{tabular}{l c|C{0.5cm}C{0.5cm}C{0.5cm}C{1cm}|C{0.5cm}C{0.5cm}C{0.5cm}C{1cm}|C{0.5cm}C{0.5cm}C{0.5cm}C{0.9cm}}
\toprule
\multirow{2}{*}{Method} & \multirow{2}{*}{Baseline} & \multicolumn{4}{c|}{40 + 40 } & \multicolumn{4}{c|}{70+10 }  & \multicolumn{4}{c}{40-20 } \\
 & & $\text{AP}$ & $\text{AP}_{50}$ & $\text{AP}_{75}$ & $CoGap$  & $\text{AP}$ & $\text{AP}_{50}$ & $\text{AP}_{75}$ &$CoGap$ & $\text{AP}$ & $\text{AP}_{50}$  & $\text{AP}_{75}$ &$CoGap$ \\
\midrule
RGR & Faster R-CNN & 35.0 & 55.7 & 37.2 & \textbf{0.6\%} & 34.6 & 54.7 & 37.4 & 2.0\% & 32.5 & 52.3 & 35.0 & 1.8\% \\
CL-DETR & Deformable DETR & 40.9 & 58.8 & 43.8 & 1.1\% & 39.6 & 56.0 & 41.2 & 1.8\% & 33.6 & 50.1 & 36.4 & 1.7\% \\
GCD & Grounding Dino & 44.7 & 61.4 & 48.7 & 1.0\% & 44.8 & 61.6 & 48.7 & 1.9\% & 42.4 & 58.0 & 46.2 & 1.6\%\\
\textbf{Ours} & YOLO-Word(X) & \textbf{52.2} & \textbf{68.7} & \textbf{57.3} & 0.8\% & \textbf{50.7} & \textbf{67.0} & \textbf{55.6} & \textbf{1.7\%} & \textbf{50.9} & \textbf{66.9} & \textbf{55.7} & \textbf{1.0\%}\\
\bottomrule
\end{tabular}
\caption{IOD results on LoCo COCO in single and multi-phase setting. All results are reproduced using publicly available code from their papers. $CoGap$ denotes the $\text{AP}$ gap compared to original COCO partition.}
\label{tab:loco_result}
\end{table*}

\begin{table*}
\centering

 	\begin{tabular}{c|cccc|ccc|cccccc}
            \toprule
            \multirow{2.3}{*}{Model} & \multirow{2.3}{*}{Pseudo Label}& \multirow{2.3}{*}{CPR} & \multirow{2.3}{*}{IKS} & \multirow{2.3}{*}{CAKD} & \multicolumn{3}{c|}{COCO(70-10)} &   \multicolumn{6}{c}{COCO(40-10)} \\
            & & & & & 1-70 & 71-80 & 1-80 & 1-40 & 41-50 & 51-60 & 61-70 & 71-80 & 1-80 \\
             \midrule
             (a)&\ding{51}&           &           &           & 49.1  &  43.4 & 48.4    & 46.5 & 33.3 & 37.9 & 56.0 & 41.2 & 44.3 \\
             (b)&\ding{51}& \ding{51} &           &           & 51.0 & 45.4  & 50.3  & 49.9 & 36.6 & 42.7 & 58.2 & 41.2 & 47.3  \\
             (c)&\ding{51}& \ding{51} & \ding{51} &           & 52.4  & 45.2 & 51.5 & 52.8 & 36.9 & 43.6 & 57.8 & 43.6 & 49.1  \\
             (d)&\ding{51}&           &           & \ding{51} & 51.6  & 45.3 & 50.8 &  53.0 & 36.9 & 42.6 & 58.0 & 43.9 & 49.2   \\
             (e)&\ding{51}& \ding{51} & \ding{51} & \ding{51} & \textbf{53.4}  & \textbf{45.4} & \textbf{52.4}   & \textbf{54.5} & \textbf{38.3} & \textbf{44.6} & \textbf{58.9}& \textbf{44.6} & \textbf{50.6} \\
             \bottomrule
	\end{tabular}
    	\caption{Investigation on the effectiveness of the main components in our method, measured in AP\%.}
        \label{tab:ablation_components}
\end{table*}

\subsection{Experiment Setup}

\textbf{Datasets and Metrics.} 
To evaluate our approach, we utilize the MS COCO 2017~\cite{lin2014microsoft} datasets with protocols established in prior studies~\cite{mo2024bridge,kim2024sddgr}. 
We also evaluate on the proposed LoCo COCO benchmark to provide more realistic assessment. 
We use standard COCO metrics: mAP across IoU thresholds 0.5-0.95, mAP@0.5, and mAP@0.75. We also report AbsGap (absolute mAP Gap) and RelGap (relative mAP Gap) compared to joint training to quantify catastrophic forgetting.

\textbf{Implementation Details.} Our method is implemented on YOLO-World (X). We use batch size 16 on 4 RTX 3090 GPUs with learning rates of $2 \times 10^{-5}$ (backbone) and $2 \times 10^{-4}$ (neck and head). Training uses an AdamW ~\cite{loshchilov2017decoupled} optimizer for 20 epochs with mosaic augmentation disabled after epoch 10. In the IKS module, the proportion of selected kernels $\mathcal{K}$ is set to 20\% during the base stage and 12\% during incremental stage.

\subsection{Comparison with State-of-the-Art Methods}

We assess our method under both single-step and multi-step IOD settings on COCO. We compare YOLO-IOD with recent state-of-the-art methods, including two-stage detectors (BPF, RGR), DETR-based approaches (CL-DETR, SDDGR~\cite{kim2024sddgr}, DCA~\cite{zhang2025dca}), and approaches built upon open-vocabulary models ( TLR~\cite{zhang2024learning}, GCD~\cite{wang2025gcd}). 
Additionally, we reproduce classic response-based distillation method ERD and recent generative replay method RGR on the YOLO-World architecture for further comparison.

\subsubsection{Single-Step Incremental Setting.}
We first evaluate performance under the 40+40 and 70+10 settings, where 40 and 10 classes are added. As shown in Tab.~\ref{tab:single_stage}, YOLO-IOD achieves consistent performance improvements over previous methods. Specifically, under the 40+40 configuration, our method surpasses the previous best approach RGR by 1.5 in AP, with AbsGap of only 1.5, and dramatically reduces the relative performance gap to joint training from 5.5\% to 2.7\%. Under the 70+10 setting, YOLO-IOD outperforms RGR by 3.3\% in AP and achieves a remarkably low RelGap of 3.9\% compared to the upper bound, while maintaining strong performance across all metrics. Note that RGR is a type of generative replay-based method, whereas our method requires no replay at all. 

\subsubsection{Multi-Step Incremental Setting.}

To better reflect real-world scenarios where new categories are continually introduced, we augment our multi-phase evaluation (40-10, 40-20) with the longer 20-20 and 10-10 settings, 
incrementally adding 20 or 10 classes per stage respectively until all 80 are learned.
These longer configurations are especially important for evaluating incremental detectors in realistic deployment scenarios.
As shown in Tab.~\ref{tab:multi_step_results}, our method consistently achieves the best performance in these long-term setups, with improvement over RGR from 3.3\% to 6.3\% AP. 
Notably, under the challenging 10-10 setting with 8 incremental phases, YOLO-IOD demonstrates exceptional performance, achieving only 8.8\% RelGap at the final stage and substantially outperforming RGR (20.3\% RelGap) and CL-DETR (48.1\% RelGap). 
The results indicate that YOLO-IOD can be successfully tailored for real-world scenarios demanding continuous adjustment to emerging object classes by adequately addressing knowledge conflicts.

\subsection{Evaluation on LoCo COCO}
We evaluate both recent methods and our YOLO-IOD on the more realistic LoCo COCO benchmark, as shown in Tab.~\ref{tab:loco_result}.
Compared to the original COCO benchmark, all methods experience an AP drop of 0.6\% to 2.0\% on LoCo COCO. which indicates the impact of data leakage in previous incremental settings. 
Despite this, YOLO-IOD consistently achieves strong performance across all incremental scenarios, demonstrating robustness even when inter-stage image overlap is removed. 
Specifically, YOLO-IOD outperforms the prior best method GCD, by 7.5, 5.9, and 8.5 AP in the 40+40, 70+10, and 40+20 settings, respectively. 
These results highlight both the practical value of YOLO-IOD for realistic continual detection and the necessity of LoCo COCO as a practical evaluation benchmark for IOD.

\begin{figure}
	\centering
	\includegraphics[width=0.85\linewidth]{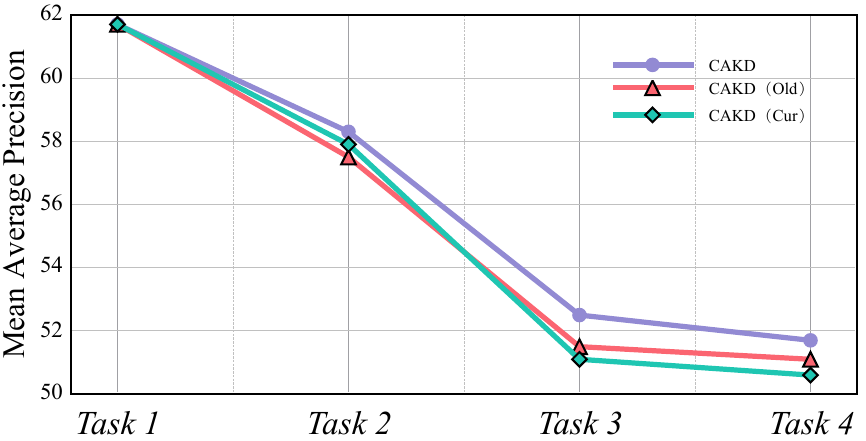}
	\caption{Ablation study within the CAKD module on the COCO 20-20 incremental setting.}
	\label{fig:CAKD_ablation_study}
\end{figure}

\subsection{Ablation Study}

\begin{figure}
	\centering
	\includegraphics[width=1\linewidth]{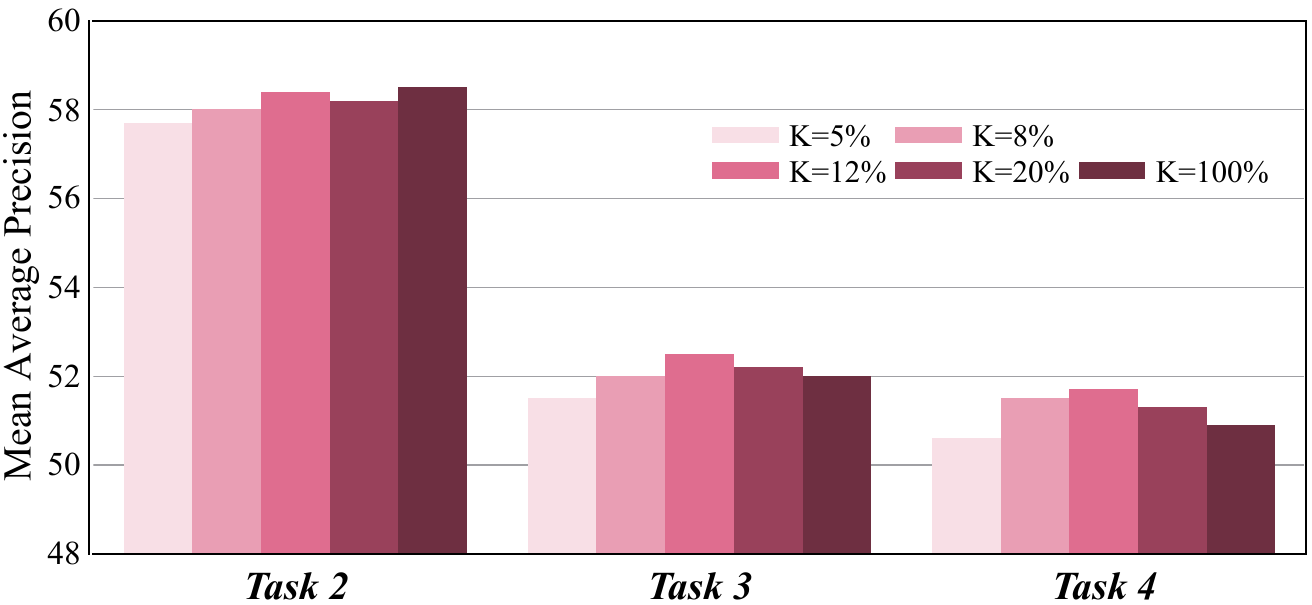}
	\caption{Ablation of kernel selection ratio $\mathcal{K}$ in IKS on COCO 20-20 incremental setting.}
	\label{fig:IPS_ablation_study}
\end{figure}

\textbf{Effectiveness of Main components.} As shown in Tab.~\ref{tab:ablation_components}, we progressively add modules to demonstrate their individual contributions in 70-10 and 40-10 settings. 
Starting from the standard pseudo-labeling baseline, which achieves 48.4\% AP on 70-10 and 44.3\% on 40-10, adding CPR significantly improves performance to 50.3\% and 47.3\%, respectively, demonstrating its effectiveness in mitigating foreground-background confusion. 
Incorporating IKS further boosts results to 51.5\% AP for 70-10 and 49.1\% for 40-10, highlighting the value of selective parameter updating. 
Notably, when CAKD is applied individually, AP rises from 48.4\% to 50.8\% in 70-10 and from 44.3\% to 49.2\% in 40-10, showcasing its strong distillation capability. 
When CAKD is combined with the previous modules, further gains can be obtained, raising AP from 51.5\% to 52.4\% in 70-10 and from 49.1\% to 50.6\% in 40-10. 
These results confirm that all three components CPR, IKS, and CAKD work synergistically, each contributing significant performance improvements and jointly addressing fundamental knowledge conflicts. Consequently, the complete YOLO-IOD framework achieves superior results.



\textbf{Ablation Study on CAKD.} As shown in Fig.~\ref{fig:CAKD_ablation_study}, we compare three CAKD variants: using only the old teacher detector, only the current teacher detector, and the full dual-teacher. In early stages, the Current-Only variant performs better by facilitating fast adaptation to new categories. As tasks accumulate, the Old-Only variant becomes more effective at preserving prior knowledge, reflecting the shift from plasticity to stability. The full CAKD consistently achieves the best results, validating the advantage of asymmetric distillation that combines both knowledge sources.

\textbf{Impact of IKS Kernel Selection Ratio $\mathcal{K}$.}
As depicted in Fig.~\ref{fig:IPS_ablation_study}, smaller ratios (e.g., 5\%) restrict the adaptation of the model, while larger ratios (e.g., 20\%) induce forgetting due to excessive updating of parameters. We observed that when setting $\mathcal{K}=12\%$ it achieves the optimal trade-off, underscoring the significance of regulating the amount of parameters to be updated in the IOD.

\section{Conclusions}

This study introduces YOLO-IOD, a novel real-time IOD framework that leverages the pretrained YOLO-World model. 
By effectively addressing the challenges of foreground-background confusion, parameter interference, and misaligned knowledge distillation through the deployment of three well-designed modules: Conflict-Aware Pseudo-Label Refinement, Importance-based Kernel Selection, and Cross-Stage Asymmetric Knowledge Distillation, YOLO-IOD achieves an optimal balance between retaining prior knowledge and acquiring the ability to detect novel category objects. Furthermore, we present LoCo COCO benchmark, which successfully mitigates data leakage and aligns category partitions with real-world co-occurrences. 
YOLO-IOD shows outstanding performance on both conventional COCO benchmark and LoCo COCO benchmark under various single-step and multiple-step settings.

\section{Acknowledgements}

This work was supported in part by the National Natural Science Foundation of China (NSFC) under Grant 62576282, 62476223; in part by the National Key Research and Development Program of China under Grant 2024YFF1306501, 2024YFB4303700; in part by Innovation Capability Support Program of Shaanxi (Program No. 2024ZC-KJXX-043); in part by the STI2030-Major Project under Grant 2022ZD0208805; in part by the Natural Science Basic Research Program of Shaanxi Province (2024JC-DXWT-07).

\bibliography{aaai2026}

\end{document}